\documentclass[10pt]{elsarticle}
\usepackage{lineno,hyperref}
\usepackage{booktabs,makecell}
\usepackage{multicol}
\usepackage{multirow}
\usepackage{amsmath,stmaryrd}
\usepackage{graphicx}
\usepackage{subfigure}
\usepackage{enumerate}
\usepackage{pgfplots}
\usepackage{tikz}
\usepackage{algorithm}
\usepackage[noend]{algpseudocode}

\usepackage{xcolor,colortbl}

\usepackage{array}
\usepackage{etex}

\journal{Journal of Pattern Recognition}

\begin{document}

\begin{frontmatter}

\title{Deep Adaptive Learning for Writer Identification based on Single Handwritten Word Images }

\author{Sheng He\corref{mycorrespondingauthor}}
\cortext[mycorrespondingauthor]{Corresponding author}
\ead{heshengxgd@gmail.com}

\author{Lambert Schomaker\corref{damma}}
\ead{L.Schomaker@ai.rug.nl}

\address{Bernoulli Institute for Mathematics, Computer Science and Artificial Intelligence, University of Groningen, 
PO Box 407, 9700 AK, Groningen, the Netherlands}

\begin{abstract}
There are two types of information {\color{blue}in} each handwritten word image: explicit information which can be easily read or derived directly, such as lexical content or word length, and implicit attributes such as the {\color{blue}author's} identity. 
{\color{blue}Whether features learned by a neural network for one task can be used for another task remains an open question}.
In this paper, we present a deep adaptive learning method for writer identification based on single-word images using multi-task learning. 
An auxiliary task is added to the training process to enforce the emergence of reusable features.
Our proposed method transfers the benefits of {\color{blue}the} learned features of a convolutional neural network from an auxiliary task such as explicit content recognition to the main task of writer identification in a single procedure. Specifically, we propose a new adaptive convolutional layer to exploit the learned deep features. 
A multi-task neural network with one or several adaptive convolutional layers is trained end-to-end, to exploit robust generic features for a specific main task, i.e., writer identification.
Three auxiliary tasks, corresponding to three explicit attributes of handwritten word images (lexical content, word length and character attributes), {\color{blue}are evaluated}.
Experimental results on two benchmark datasets show that the proposed deep adaptive learning method can improve the performance  {\color{blue}of} writer identification based on single-word images, {\color{blue}compared} to non-adaptive and simple linear-adaptive approaches.
\end{abstract}

\begin{keyword}
Writer identification,
Deep adaptive learning,
Handwritten word attributes,
Multi-task learning
\end{keyword}

\end{frontmatter}

\section{Introduction}

Writer identification is a typical pattern-recognition problem {\color{blue}which aims to recognize the author of a handwritten passage from an image of it}. 
The authorship is an implicit (indirect) {\color{blue}attribute of a handwritten document}.
A writer-identification system usually extracts the handwriting-style information from the query document image and compares it with the style information of known writers. 
The handwriting style is usually measured by {\color{blue}a number of} geometric features, such as global statistics of ink traces~\cite{bulacu2007text,brink2012writer} or the distribution of graphemes~\cite{siddiqi2010text,he2015junction}.
The reliability of a typical writer-identification system using handcrafted features depends on the amount of text in handwritten images. 
{\color{blue}In~\cite{brink2008much} it was found that when using traditional writer identification approaches, about 100 letters are needed per sample of Western handwriting to achieve the very satisfactory results.}

However, in the digital era, {\color{blue}handwriting is an increasingly rare activity.}
In forensic applications, this requires a new approach {\color{blue}to be} able to recognize the writer based on {\color{blue}the} very small amount of {\color{blue}available text, which may be as little as a single word}. 
In this paper, we study the writer-identification problem based on single-word images, which is a challenging problem because the information contained in a single word is a highly limited information source for modelling {\color{blue} an author's writing style}.
In order to solve this problem, the convolutional neural network (CNN)~\cite{NIPS2012_4824} is used for writer identification in this paper because it can learn discriminative and hierarchical features {\color{blue}at} different abstraction levels from raw data and it has achieved {\color{blue}good} performance on various applications in computer vision~\cite{NIPS2012_4824,zhang2016cosaliency,girshick2015fast} and handwriting recognition~\cite{sudholt2016phocnet,zhang2017online}.

\begin{figure}
    \centering
    \includegraphics[width=0.95\textwidth]{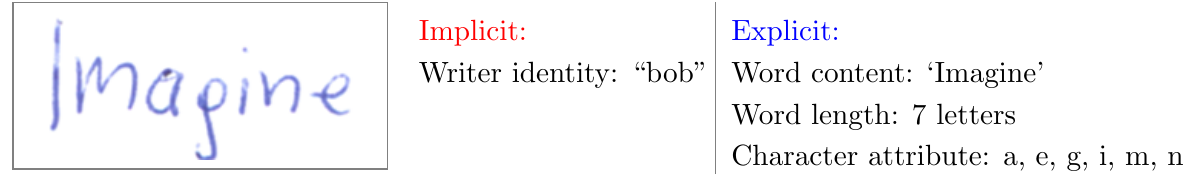}
    \caption{There are two types of information {\color{blue}in any given image of a handwritten word:} implicit information, such as the writer and explicit information such as the exact word content, word length and character presence.}
    \label{fig:wordinfo}
\end{figure}

There are two types of information {\color{blue}in any given image of a handwritten word:}
explicit information, such as the lexical content, word length and character attributes, and implicit information, such as the {\color{blue}writer's} identity. Explicit information can be derived relatively easily from the image sample itself, whereas implicit information must be derived from a separate source. An example is shown in Fig.~\ref{fig:wordinfo}.
The derivation or estimation of implicit and explicit information actually corresponds to different tasks, such as word recognition and writer identification, which would be treated separately in traditional pattern recognition {\color{blue}methods}.
{\color{blue}Word recognition methods} extract shape features which come from a sequence of curvilinear strokes in word images~\cite{dasgupta2016holistic}, while writer-identification methods extract the slant, curvature or ink-width distribution to capture the writing style {\color{blue}applied to form the handwritten word}~\cite{bulacu2007text,brink2012writer}. 
This distinction appears to {\color{blue}involve} a loss of resources and a lack of generalizability, which {\color{blue}becomes clearer as more tasks are attempted -} such as document dating or historical writing-style classification {\color{blue}- for which  completely new approaches need to be designed}. At the same time, {\color{blue}specific aspects of shape information can be expected to be useful for more than one task.}

Performing more than one task on the same input data corresponds to the multi-task learning problem~\cite{zhang2014facial,zhang2012robust,hwang2016self} {\color{blue}and this has been achieved successfully} in many applications. In this paper, we {\color{blue}apply} multi-task learning {\color{blue}to} the same input to train neural networks on writer identification with an additional auxiliary task, i.e., word-text recognition, which addresses {\color{blue}the} explicit information present in a handwritten word image.

It has been shown in~\cite{yosinski2014transferable} that {\color{blue}the layers of learned convolutional neural networks transition from being more general, towards the input layer, to being more task-specific, towards the output layer.}
The layers close to the input will contain more general representations which can be shared between different tasks in multi-task learning. However, layers close to the output become more specific to each task and they cannot be used directly for other goals than the one {\color{blue}trained for}.
In the literature, transfer learning is usually adopted to transfer general features between multiple tasks by sharing several lower layers closer to the input. Adaptive learning can be applied to transfer the specific features of the auxiliary task to the main task by a linear combination of input activation maps, {\color{blue}in order to achieve better performance in the main task~\cite{misra2016cross,ruder2017sluice}}.

Because the information capacity of the convolutional neural network is quite large, as expressed in the number of weights, it is to be expected that it can learn different features for different tasks. For example, the features learned for word recognition might capture word-shape information, while features learned for writer identification might capture the ink density or curvature information in the handwritten images.
Deep adaptive learning aims to transfer and mix the learned features from one task {\color{blue}to and} with another in order to improve performance {\color{blue}by} using an integral end-to-end training procedure.
 This is expected to work due to the following two reasons:
(1) A deep neural network that is trained just for the writer identification task {\color{blue}might} be overfitted {\color{blue}for} the writer identification problem and therefore {\color{blue}it is possible} that it does not generalize well within this task. Conversely, adapting the trained features to an additional task during {\color{blue}the} training itself is assumed to introduce a regularization which can reduce the risk of over-fitting~\cite{ruder2017overview} and improve the performance on unseen data.
(2) Transferring the learned features from other tasks can be considered {\color{blue}to be} feature combination over different pathways in a particular layer. Feature combination has been shown to provide better performance~\cite{bulacu2007text,he2017beyond}.

In this paper, we will apply deep adaptive learning to the application of writer identification under the difficult condition of a very small sample, {\color{blue}for instance an} isolated word. This is a highly challenging problem because the writer-related style information will be very limited in the {\color{blue}small} word image.
We {\color{blue}will} choose different attributes of handwritten word images as the auxiliary task to demonstrate the effectiveness of the proposed deep adaptive learning method.  In particular, we {\color{blue}will} choose three tasks as the auxiliary task in multi-task learning: word recognition, word length estimation and character attribute recognition.
When showing a word image to a human reader, the word content will be recognized first, but we can ask additional questions about word length or about {\color{blue}the} shape attributes of the characters it contains. 
In fact, there may be several other explicit pieces of information when we read a handwritten word image, such as the stroke width of the ink caused by the writing instruments, or the number of circle and cross line intersections present in the word image, etc. To test the hypothesis that the proposed deep adaptive learning method works, we selected explicit information which is very easy to derive (word label, word length, number of letters), {\color{blue}and does not require} additional complicated pattern-recognition tools such as a circle detector.
In general, the auxiliary tasks should not introduce expensive additional labelling in a real-world application.

The contributions of this paper are summarized as follows: (1) We study the writer-identification problem based on single word images, which is a very challenging real-life application problem.
(2) We propose a non-linear deep adaptive learning method to transfer the features learned from an auxiliary task to the writer-identification task, fully integrated within the training procedure. We will demonstrate that  the proposed deep adaptive learning {\color{blue}method} will provide better performance than non-adaptive or linear-adaptive learning methods.
(3) We evaluate three different auxiliary tasks for writer identification (word recognition, word-length estimation and character-attribute recognition), which all improve the performance {\color{blue}to different degrees}. 

Signature identification or verification aims to verify the individual's identity {\color{blue}from} handwritten signatures~\cite{impedovo2008automatic}. 
The problem of writer identification based on single-word images is somewhat similar to {\color{blue}the} signature identification problem, since both extract an individual's writing style. 
However, writer identification based on single-word images aims to {\color{blue}identify} the writer based on any given word, as opposed to the signature, which is stable {\color{blue} to the individual} and usually designed by {\color{blue}that} person to have a unique personal shape, unlike isolated handwritten words from a normal piece of text.
Our proposed method attempts to model the general writing style from a set of isolated handwritten word images in the training set.
		
This paper is organized as follows. In Section~\ref{sec:relatedwork} we provide an overview of related work. We introduce the proposed adaptive learning in Section~\ref{sec:approach}. The experimental results are presented {\color{blue}and discussed} in Section~\ref{sec:experiments}.
{\color{blue}The last section concludes the paper}.

\section{Related Work}
\label{sec:relatedwork}

\begin{table*}
    \centering
    \caption{Advantages and disadvantages of different writer-identification methods.}
    \label{tab:procons}
    \resizebox{0.95\textwidth}{!}{
    \begin{tabular}{p{1.5cm}|p{5cm}|p{4cm}|p{4cm}}
            \hline\hline 
            References & Features & Advantages & Disadvantages \\
            \hline
            \multicolumn{4}{c}{\textbf{Texture-based features}} \\
            \hline
          \cite{bertolini2013texture,niorlaou2015sparse,djeddi2013text} & Each pixel is described by local binary patterns (LBP and LPQ) and the feature from the whole text-block is computed by a normalized histogram. & Easy to compute without binarization and segmentation.
          Parameter-free methods. & The LBP histogram itself is not effective and some post-processing steps are usually applied, such as GLCM, PCA or Run-length.  \\
          \hline
          \cite{said2000personal,helli2010text,newell2014writer} & {\color{blue}Computes} the response of handcrafted Gabor-based filters to describe the texture properties of handwriting style. & Each type of filter captures certain {\color{blue}handwritten character} shapes, thus the feature is easy to understand and explain to end users. & Requires careful design or selection of the parameter values of filters.  \\
     	  \hline 
     	  \cite{bulacu2007text,brink2012writer,he2014delta,he2016co} & {\color{blue}Extracts} the writing style information based on ink trace by edge or contour angles. The feature vector is the joint distribution of angles on each position of ink trace. & Fast and efficient to compute. Captures curvature and slant information of the writing style.  & Requires  binarization or high-contrast images.  \\
     	  \hline 
     	  \multicolumn{4}{c}{\textbf{Grapheme-based features}} \\
     	  \hline 
    	  \cite{schomaker2004automatic,schomaker2007using,ghiasi2013offline} & {\color{blue}Computes} contour and stroke fraglets {\color{blue}for} handwritten characters.  & Informative and each grapheme represents {\color{blue}an entire} letter or parts of letters which are shared between different characters. & Requires binarization, segmentation and an effective fragmentation heuristic for connected-cursive handwritten documents.\\
    	  \hline 
		  \cite{siddiqi2010text,djeddi2013codebook} & {\color{blue}Extracts} small patches on handwritten characters. & The patches are small so that they can be used for many different scripts and can be generated randomly.
		 & {\color{blue}No pattern in the small patches carries} any semantic information. The patches are too small and the distribution is not distinctive enough for graphemic style differences, thus performance is limited. \\ 
    	  \hline 
    	  \cite{abdi2015model} & {\color{blue}Uses} the elliptic model to generate an exhaustive number of graphemes. & Model-driven method without {\color{blue}codebook training} (grapheme selection involved to obtain a compact feature vector).  & Morphological operations are needed to match the handwriting contours and graphemes. {\color{blue}Due to elliptic model limitation}, it is only evaluated {\color{blue}for} Arabic texts.\\
    	  \hline 
    	  \cite{he2015junction} & {\color{blue}Extracts} junction parts on the ink traces. & Junctions are prevalent in different handwritten scripts. Their shape contains the writing style of the author and can be used for cross-script writer identification. & Requires binarized images and the performance is limited in poor-quality images.\\
          \hline\hline
    \end{tabular}
}
\end{table*}

Most writer identification methods are text-independent,  {\color{blue}extracting} features from large image regions - {\color{blue}such as pages}, text blocks or sentences - instead of small word images.
In the last few decades, many {\color{blue}specially} handcrafted features {\color{blue}have been} designed to extract low-level features from handwritten images. {\color{blue}These} can be roughly grouped into two groups:  textural-based and grapheme-based features.

Textural-based methods extract statistical information from the entire text blocks as features. 
Considering the handwritten text as a texture, textural features are extracted to measure the similarity {\color{blue}in handwriting style} between different handwritten document images. Local binary patterns (LBP)~\cite{bertolini2013texture,niorlaou2015sparse} and local phase quantization (LPQ) are proposed in~\cite{bertolini2013texture} and the run-length of LBP is proposed in~\cite{he2017writer} for writer identification. 
A run-length histogram {\color{blue}with} four principal directions is proposed in~\cite{djeddi2013text} for writer identification in a multi-script environment. 
Filter-based features, such as Gabor~\cite{said2000personal}, XGabor~\cite{helli2010text} and oriented Basic Image Feature Columns (oBIF Columns)~\cite{newell2014writer}, {\color{blue}have also been studied}.
Some features {\color{blue}can be} extracted from the contours of the ink trace, such as Hinge-based features~\cite{bulacu2007text,brink2012writer,he2014delta,he2016co}, which extract the slant property of characters {\color{blue}alongside} other information, such as stroke width~\cite{brink2012writer} and curvature information~\cite{siddiqi2010text}. 
Other features, such as symbolic representation~\cite{alaei2014new} and k-adjacent segments (kAS)~\cite{jain2011offline,jain2014combining} are also used for writer identification.
Gaussian Mixture Models (GMMs) {\color{blue}are used} to model a {\color{blue}person's}
handwriting in~\cite{schlapbach2006off} and Hidden Markov Model (HMM)-based recognizers are used in~\cite{schlapbach2007writer}.

Grapheme-based features extract allographic patterns and map them into a common space (also known as a codebook). 
Connected-component contours (CO$^3$) are proposed in ~\cite{schomaker2004automatic,ghiasi2013offline} for writer identification {\color{blue}using} upper-case Western scripts, and {\color{blue}have more recently} been extended to Fraglets~\cite{bulacu2007text,schomaker2007using} {\color{blue}for} cursive handwriting documents.
Small patches extracted from characters are used as graphemes in~\cite{siddiqi2010text,djeddi2013codebook} and synthetic graphemes {\color{blue}which have been} generated based on the beta-elliptic model are used in~\cite{abdi2015model} on Arabic handwritten document images.
The junctions {\color{blue}in} handwritten images are very useful for measuring the handwriting style and they are considered {\color{blue}to be} basic elements of the handwritten text for writer identification in~\cite{he2015junction}.
SIFT feature~\cite{lowe2004distinctive} and
RootSIFT descriptor are also used for writer identification~\cite{wu2014offline,christlein2017writer}.
Both the textural-based and grapheme-based features can be used to generate more powerful features by the co-occurrence or joint feature principle, which can be found in~\cite{he2017beyond}.
Table~\ref{tab:procons} shows {\color{blue}the} advantages and disadvantages of each method mentioned above.

 Writer identification based on scarce data has also been investigated. For example, Alaei and Roy~\cite{alaei2014new} propose a writer identification method based on {\color{blue}the} line and page-level, {\color{blue}where} performance {\color{blue}at} the page-level is higher than the performance {\color{blue}at} the line-level.
Similar conclusion were obtained in~\cite{newell2014writer}, {\color{blue}where} comparable performance was achieved based on at least three lines using the oBIF features with delta encoding. 
Adak and Chaudhuri~\cite{adak2015writer} propose a writer identification method {\color{blue}for} isolated Bangla characters and numerals.
The handcrafted features usually need more text because statistical information is used to model the writing style, and the corresponding feature distribution must be stable and representative when more texts are given.
However, there are usually only a few letters/characters in single-word images. Therefore, the handcrafted feature distribution extracted on their basis does not approximate the true distribution of the writing style, {\color{blue}resulting in poor} performance. If the amount of text is limited, the importance of small structural {\color{blue}fragments} of shape evidence becomes {\color{blue}greater}. We expect convolutional deep learning to be able to learn the necessary feature-kernel shapes.

Recently, deep learning {\color{blue}has also been} used for writer identification. For example, a neural network can be trained based on a small block, segmented from the text line with a sliding window~\cite{fiel2015writer} or a texture block~\cite{tang2016text}.
A deep multi-stream CNN is proposed in~\cite{xing2016deepwriter} to learn deep features for writer identification.
As mentioned above, a deep neural network can learn  discriminative and hierarchical features~\cite{zeiler2014visualizing} and can recognize writers {\color{blue}on the basis of} less data.
Therefore, deep learning can capture a writing style based on single-word images. 
However, all of these methods consider writer identification as a single task. Multi-task learning aims to jointly learn classifiers for {\color{blue}several} related tasks using shared representation. For example, the method proposed in~\cite{vezhnevets2010towards} uses an external task to improve semantic segmentation in natural images.
Other multiple-task learning methods using CNN include edge labels and surface normals~\cite{wang2015designing} and face detection and face landmark detection~\cite{zhang2014facial}.
Hwang and Kim~\cite{hwang2016self} propose multi-task learning for the classification and localization {\color{blue}of} medical images.
Misra et al.~\cite{misra2016cross} propose a \textit{cross-stitch unit} {\color{blue}in order} to learn an optimal combination
of shared and task-specific representations among multiple tasks.
Multi-task learning is also evaluated in natural language processing, which demonstrates that adding an auxiliary task can help improve the performance of the main task~\cite{bjerva2017will}.
Our proposed method uses a non-linear adaptive strategy which introduces a convolutional layer to transfer features from the auxiliary task to the main task.

\section{Proposed Method}
\label{sec:approach}
In this section, we describe the proposed method for writer identification based on single-word images using deep adaptive learning. We first introduce the structure of the CNN used for {\color{blue}the} multi-task learning, with the writer identification as {\color{blue}its} main task. After that, we show how to transfer and adapt the learned features from the auxiliary task to the main task to improve the performance of writer identification.

\subsection{Main Architecture of the Convolutional Neural Network}

\begin{figure*}
	\centering
	\includegraphics[width=\textwidth]{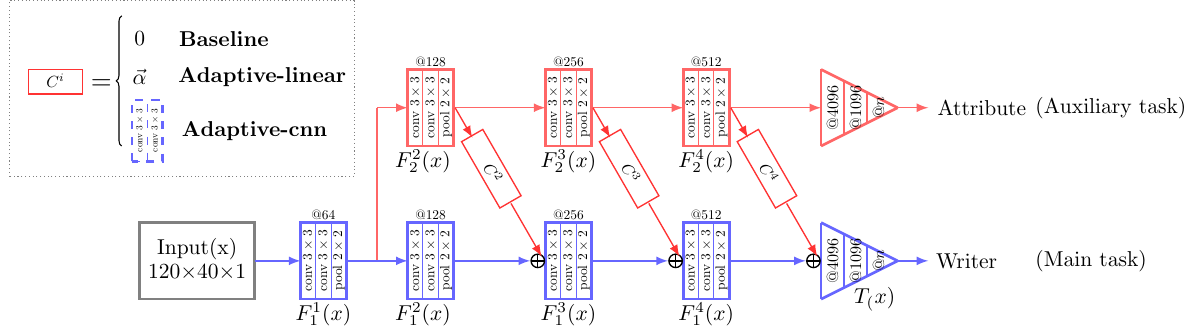}
        \caption{Overall {\color{blue}diagram} of the proposed deep adaptive convolutional neural network. The input is a grayscale word image of $120\times40$ pixels.  There are eight convolutional layers {\color{blue}for each task}, four max-pooling layers and three fully-connected layers in this framework. Each convolutional block contains two convolutional layers and one max-pooling layer (the kernel size is denoted in the boxes). $F_1^i(x)$ denotes the feature maps on the $i$-th block for the main writer identification task (blue) and $F_2^i(x)$ for the auxiliary task (red). 
The notation @`$k$' above each block indicates the number of kernels used in the convolution. The number $n$ in the last layer represents the number of classes.
The block $C^i(\cdot)$ is an adaptive function, which has three types in this paper:  \textbf{Baseline} when  $C^i(\cdot)=0$, \textbf{Linear-adaptive} when  $C^i(\cdot)=\vec{\alpha}$ and \textbf{Deep-adaptive} when  $C^i(\cdot)=cnn$, i.e., a deep network itself.}
    \label{fig:adaptivenet}
\end{figure*}

The architecture of our convolutional neural network is a multi-task adaptation of the AlexNet structure~\cite{NIPS2012_4824}, which is shown in Fig.~\ref{fig:adaptivenet}.
The architecture contains a pathway for the main task and a pathway for the auxiliary task.
The two pathways interact at several possible layers where {\color{blue}adaptation} takes place.
For the main task, the pathway consists of {\color{blue}eight} convolutional layers, {\color{blue}with four} max-pooling layers after every two convolutional layers in order to increase the depth of network and {\color{blue}three} fully connected layers.
All {\color{blue}of the inputted} handwritten word grayscale images are resized to 120$\times$40$\times$1. 
The size of {\color{blue}the} receptive field is 3$\times$3 {\color{blue}for all of the convolutional layers}, which is widely used in deep neural networks~\cite{simonyan2014very,he2016deep}. The convolutional stride is fixed {\color{blue}at} one pixel {\color{blue}for all of the} convolutional layers. The number of filters of each convolutional layer is {\color{blue}depicted} in Fig.~\ref{fig:adaptivenet}.
The first two convolutional layers {\color{blue}are} shared by both task pathways. 
For the auxiliary task, each layer mirrors a corresponding layer in the pathway for the main task. Details concerning this configuration {\color{blue}are presented below}.

After each convolutional or fully-connected layer (except {\color{blue}for} the last softmax layer), the leaky-ReLU (Rectified linear unit) activation function~\cite{maas2013rectifier} is used to avoid neurons {\color{blue}dying} if their input activations {\color{blue}are} below the threshold, which is defined as:
$f(x) = \max(\lambda x,x)$ (in this paper, $\lambda=0.1$).
Spatial pooling is also very important in CNN models to integrate {\color{blue}the} available information and {\color{blue}simultaneously} to reduce the size of {\color{blue}the} feature maps. In our model, a max-pooling layer with a kernel size {\color{blue}of} 2$\times$2 and a stride step {\color{blue}of} 2 is implemented after every two convolutional layers (see Fig.~\ref{fig:adaptivenet}) to reduce the size of the input representation.
Dropout layers~\cite{srivastava2014dropout} {\color{blue}a} dropout rate {\color{blue}of} 50\% are applied after the first two fully connected layers in order to mitigate the over-fitting problem. The last layer is usually a softmax layer for single label recognition. 
For the loss function, we {\color{blue}applied} the cross-entropy loss, which measures the dissimilarity between the true label distribution and the predicted label distribution.

\subsection{Auxiliary pathway and adaptive transfer}
As shown in Fig.~\ref{fig:adaptivenet}, the auxiliary pathway receives shared-feature patterns and {\color{blue}the} layers are organized in parallel to the main pathway.
It would be beneficial to adapt the learned high-level task-specific features from the layers near the output layer of the neural network of the auxiliary task to the main task {\color{blue}in order} to improve the performance, if the learned features from layers near the output layer are reusable in another task~\cite{yosinski2014transferable}. 
However, it is unlikely that the learned features can be used {\color{blue}as they are, and} some task-specific fine-tuning is likely to be required.
Therefore, we propose an adaptive network  which transfers the representation from layers near the output layer of an auxiliary task to the main task via an adaptive function, $C^i(\cdot)$.
Given two activation maps $r(F_1^i)$ and $r(F_2^i)$ from the convolutional layer $F^i$ ($i=2,3,4$ in Fig.~\ref{fig:adaptivenet}) for both tasks ($F_1$ for writer identification and $F_2$ for the auxiliary task), we learn a combination of $r(F_1^i)$ and $r(F_2^i)$ and feed it as input to the next layer $F_1^{i+1}$ of the main task by:
\begin{equation}
\label{eq:res}
in(F_1^{i+1}) = r(F_1^i) + C^i\Big(r(F_2^i)\Big)
\end{equation}
where $in(F_1^{i+1})$ is the input of the next layer $F_1^{i+1}$ and $C^i(\cdot)$ is an adaptive function on the layer $F_2^i$ which adapts the representation $r(F_2^i)$ from the auxiliary task to the main task of writer identification.

Different adaptive functions $C^i(\cdot)$ can be applied, and in this paper, we evaluate three types of functions as follows:
\begin{enumerate}
	\item \textbf{Baseline} ($C^i(\cdot)=0$): The adaptive function is zero, which means that there is no {\color{blue}adaptation} between two tasks. This can be considered as the baseline, in which two tasks share the first two convolutional layers without adaptation.
	\item \textbf{Linear-adaptive} ($C^i(\cdot)=\vec{\alpha}$): The adaptive function is a linear mix function, which is similar {\color{blue}to} the cross-stitch unit proposed in~\cite{misra2016cross}. In this case, Eq.~\ref{eq:res} becomes:
	 \begin{equation}
	in_j(F_1^{i+1}) = \alpha_j \cdot r_j(F_1^i) + (1-\alpha_j)\cdot r_j(F_2^i)
	\end{equation} 
	where $j$ is the index of the number of activation maps in the layer $F^i$, $\alpha_j$ is the parameter which weights the activation from the main task and $1-\alpha_j$ weights the activation from the auxiliary task. Note that we set different $\alpha$ to different activation maps and the dimensionality of the $\vec{\alpha}$ vector is the same as the depth of the layer $r(F_2^i)$, i.e., the number of filters in the layer $r(F_2^i)$. Given the initialization ($\alpha$=0.5), the $\vec{\alpha}$ is also learned during training and the network can find the optimal weights of the adaptive function between the activation maps of the auxiliary and {\color{blue}the} main tasks. 
	\item \textbf{Deep-adaptive} ($C^i(\cdot)=$CNN): In this case, the adaptive function is a convolutional neural network itself. In this paper, we use two convolutional layers with the kernel 3$\times$3 and the number of kernels of each $C^i(\cdot)$ is the same as the corresponding layers $F_1^i$ and $F_2^i$ in order to make the dimension equal for the add operation. From Eq.~\ref{eq:res} we can obtain:
	\begin{equation}
	\label{eq:resres}
	C^i\Big(r(F_2^i)\Big) = in(F_1^{i+1}) - r(F_1^i)
	\end{equation}
	where $r(F_1^i)$ is the features on the $i$-th layer and $in(F_1^{i+1})$ is the input features of the $(i+1)$-th layer of the main task.  Therefore, $C^i\Big(r(F_2^i)\Big)$ is the residual features of the main task learned from layer $F_2^i$ of the auxiliary task.	
	Using the convolutional layers as the adaptive function {\color{blue}makes it possible to} capture more complex {\color{blue}structures} between the activation maps of {\color{blue}the} different tasks and find the best adaptive representations between two different tasks. These adaptive layers are also learned jointly during {\color{blue}the} training, and the loss of the main task is back-propagated through these adaptive layers.
	
\end{enumerate}

\subsection{Training}
There are two losses {\color{blue}in our network}: $\textbf{Loss}_{au}$ for the auxiliary task and $\textbf{Loss}_{wi}$ for the {\color{blue}writer-identification task}.
The cross-entropy loss function is computed in this paper for both {\color{blue}the} auxiliary and {\color{blue}the} main tasks.
The network is trained jointly for the auxiliary and writer-identification task, based on a weighting strategy in our paper.
The objective function is defined as:
\begin{equation}
\label{eq:lambda}
\textbf{Loss}_{total}= (1-\lambda) \textbf{Loss}_{au}+\lambda \textbf{Loss}_{wi}
\end{equation}
where $\lambda$ is the trade-off weight of the two losses.
At the beginning of training, these two losses are equal, so we set $\lambda=0.5$. 
In practice, we have found that the loss of the auxiliary task, which recognizes the explicit information, decreases faster than the loss of {\color{blue}the writer-identification task}. Therefore, we increase the $\lambda$ after {\color{blue}a given} iteration to fine-tune the network {\color{blue}for} writer identification.
As explained in~\cite{hwang2016self}, the relative importance of {\color{blue}the} two losses weighted by $\lambda$ can be back-propagated to the adaptive layers $C^i(\cdot)$.

\section{Experiments}
\label{sec:experiments}
In this section, we conduct experiments on two benchmark datasets for writer identification based on single-word images with three different auxiliary tasks.

\subsection{Datasets}

We evaluate our proposed methods {\color{blue}through the use of} two publicly available \textbf{CVL} and \textbf{IAM} datasets which {\color{blue}present} segmented word images with labels {\color{blue}for} both word and writer.
The proposed method is evaluated through using these two datasets separately, because {\color{blue}the} writers from these two datasets {\color{blue}differ}.

\textbf{CVL}~\cite{kleber2013cvl} consists of 310 writers, each of {\color{blue}which contributing} at least five pages in English and German. 
The word regions were {\color{blue}automatically} labelled and were evaluated by two students independently. 
{\color{blue}In order to train the network for this paper}, we select word images {\color{blue}with at least twenty instances}. {\color{blue}Ultimately, this yielded} 99,513 selected word images which were randomly split into training (70,778 word images) and testing (28,735 word images) sets.

\textbf{IAM}~\cite{marti2002iam} consists of 657 writers, each {\color{blue}contributing} at least one page in English. {\color{blue}Like} the \textbf{CVL} dataset, {\color{blue}the} word images {\color{blue}were} also provided in the dataset with labels for both word and writer. {\color{blue}Again,} we selected words {\color{blue}with} more than twenty instances, {\color{blue}yielding a total of} 49,625 images randomly split into training (35,421 word images) and testing (14,201 word images) sets.

\subsection{Implementation details}

The neural network {\color{blue}was first} initialized {\color{blue}using} the Xavier method proposed in~\cite{glorot2010understanding}, which has proven to work very well in practice and can speed up training.
{\color{blue}The} adaptive learning rate algorithm Adam proposed in~\cite{kingma2015adam} {\color{blue}was} used to train the neural network, with an initial learning rate {\color{blue}of} 0.0001.
The size of {\color{blue}the} mini-batch {\color{blue}was} set {\color{blue}to} 100 and the number of training iterations {\color{blue}was} set to 40,000.

During training, the parameter of $\lambda$ in Eq.~\ref{eq:lambda} {\color{blue}was} set to 0.5 {\color{blue}for} the first 10,000 iterations.
{\color{blue}It was then increased} by 0.066 at every 5,000 iterations, {\color{blue}up} to 0.9 at the end of training.
Our network {\color{blue}was} trained {\color{blue}using} the Tensorflow platform~\cite{abadi2016tensorflow}. 
Training {\color{blue}took} about 7.5 hours for the \textbf{Baseline} and \textbf{Linear-adaptive} CNN models and 8.5 hours for the \textbf{Deep-adaptive} model, on a single GPU (NVIDIA GTX 960 with 4G memory).

\subsection{Performance of writer identification with word recognition as auxiliary task}
\begin{figure}[!t]
	\centering
	\includegraphics[width=0.95\textwidth]{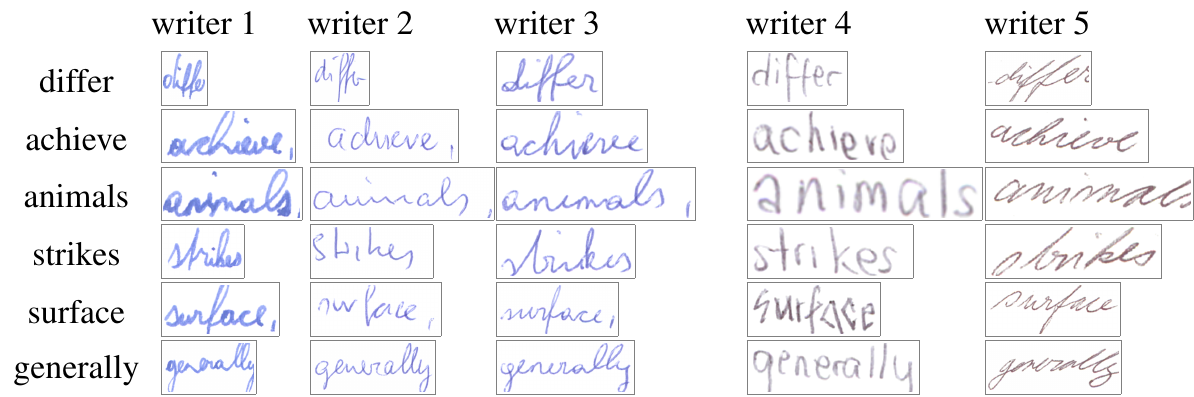}
	\caption{Examples of handwritten word images from the \textbf{CVL} dataset with different words and writers. Each image has two attributes: lexical content and {\color{blue}the writer's identity}. }
	\label{fig:wordattrimg}
\end{figure}

The lexical content of the word image is a very important information, which corresponds to the word recognition or spotting problem~\cite{van2008handwritten,almazan2014segmentation}.
{\color{blue}This section reports the experimental results} with word recognition as the auxiliary task to improve the performance of writer identification based on single-word images. Three hundred different words {\color{blue}were selected from} the \textbf{CVL} dataset and 446 different words {\color{blue}from} the \textbf{IAM} data set.
Fig.~\ref{fig:wordattrimg} {\color{blue}presents} an example of {\color{blue}the} word images with two attributes: writer and lexical content. 

Table~\ref{tab:wordrec} shows the performance of writer identification with word recognition as the auxiliary task. 
From the table we can see that the {\color{blue}word-recognition accuracies} are higher than {\color{blue}those of} writer identification, which demonstrates that writer identification (implicit information) based on single-word images is more challenging than word recognition (explicit information).
In addition, adaptive learning methods provide better results than {\color{blue}the} baseline for writer identification and the
\textbf{Deep-adaptive} model achieves the best performance on the two datasets, {\color{blue}outperforming} the \textbf{Baseline} and \textbf{Linear-adaptive} {\color{blue}models} by 3.3\% and 1.6\% on \textbf{CVL} and 3.8\% and 1.5\% on \textbf{IAM} in terms of the Top-1 recognition rate.

\begin{table}[!t]
	\centering 
	\caption{Performance of writer identification using different adaptive learning methods with \textbf{word recognition} as the auxiliary task on the \textbf{CVL} and \textbf{IAM} datasets.}
	\label{tab:wordrec}
	\resizebox{0.95\textwidth}{!}{
		\begin{tabular}{l|cc|cc|cc|cc}
			\hline\hline 
			\multirow{3}{*}{Model} & \multicolumn{4}{c|}{Writer Identification} & \multicolumn{4}{c}{Word Recognition (\textit{aux.})} \\
			\cline{2-9}
			 & \multicolumn{2}{c|}{\textbf{CVL}} & \multicolumn{2}{c|}{\textbf{IAM}} & \multicolumn{2}{c|}{\textbf{CVL}} & \multicolumn{2}{c}{\textbf{IAM}}\\
			\cline{2-9}
			& Top1 & Top5 & Top-1 & Top-5 & Top-1 & Top-5 & Top-1 & Top-5\\
			\hline 
			Baseline        & 75.3 & 92.4 & 65.7 & 83.5 & 95.1 & 99.1 & 93.5 & 98.7\\
			Linear-adaptive & 77.0 & 93.1 & 68.0 & 84.7 & 94.1 & 98.9 & 91.3 & 98.1\\
			Deep-adaptive    & \textbf{78.6} & \textbf{93.7} & \textbf{69.5} & \textbf{86.1} & 94.5 & 99.0 & 92.6 & 98.4\\
			\hline\hline
		\end{tabular}
	}
\end{table}

\begin{figure}[!t]
	\centering
	\includegraphics[width=0.95\textwidth]{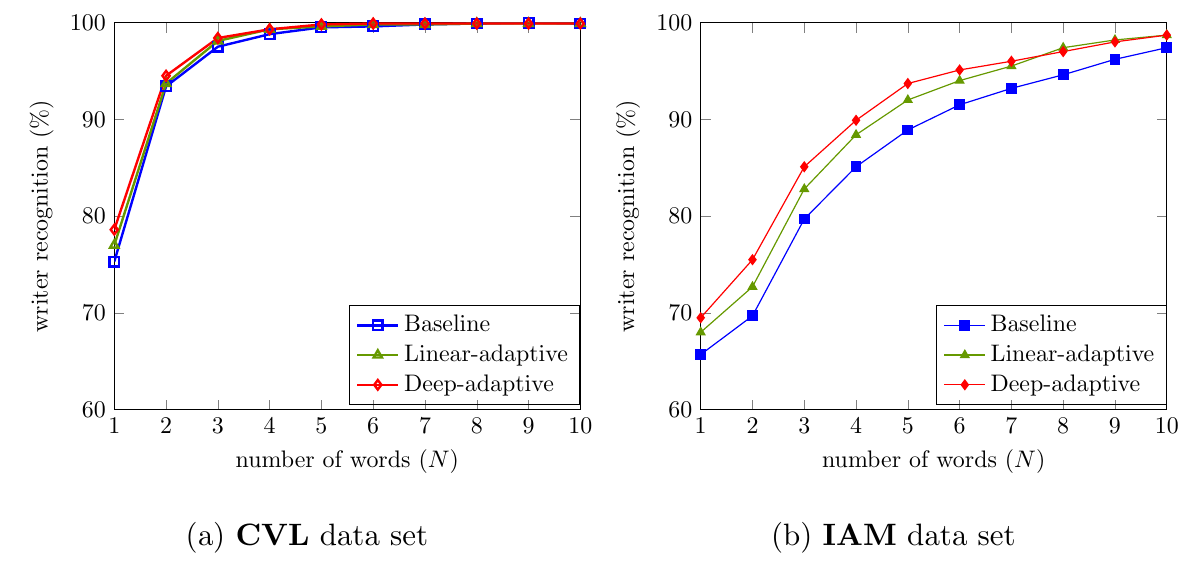}
	\caption{Performance (Top1) of writer identification {\color{blue}using} different numbers of words (from 1 to 10 words), using CNN models trained with word recognition as the auxiliary task on the \textbf{CVL} dataset (Figure (a)) and the \textbf{IAM} dataset (Figure (b)).}
	\label{fig:wordrec}
\end{figure}

Since the writer-identification performance based on single-word images is lower than {\color{blue}that of} word recognition,
another interesting question is raised: how many words are needed to achieve a higher performance for writer identification, similar {\color{blue}to the performance for} word recognition? To answer this question, we did another {\color{blue}set of experiments} about writer identification based on $N$ word images from the same writer. We randomly selected $N$ word images {\color{blue}for each writer} and put them into the trained CNN model. The average response of the last layer of the CNN model from all $N$ word images {\color{blue}was} used to recognize the writer by:
\begin{equation}
y= \frac{1}{N}\sum_i^{N}\text{CNN}(x_i)
\end{equation}
where $x_i$ is the $i$-th input image and $\text{CNN}(x_i)$ is the response of the last layer of the CNN model. The procedure {\color{blue}was} repeated 20 times for each writer and the average results {\color{blue}for different values of $N$ are} reported in Fig.~\ref{fig:wordrec}.

From Fig.~\ref{fig:wordrec} we can see that writer-identification performance {\color{blue}increases with} more word images from the same writer.  
The \textbf{Deep-adaptive} model achieves the best results with different numbers of words for writer identification.
The Top-1 performance {\color{blue}for} writer identification {\color{blue}using} the \textbf{Deep-adaptive} model was 79.1\% and 68.3\% when using one word, and {\color{blue}this} increases to 99.8\% and 92.0\% when using five words on \textbf{CVL} and \textbf{IAM}, respectively.
For the specialized textural features such as the Hinge~\cite{bulacu2007text}, the minimum text for writer identification is 100 characters~\cite{brink2008much}. However, the write-identification performance using CNN models with five words are comparable to the {\color{blue}results obtained for} textural features.

\subsection{Performance of writer identification with word length estimation as auxiliary task}
\begin{figure}[!t]
	\centering
	\includegraphics[width=0.95\textwidth]{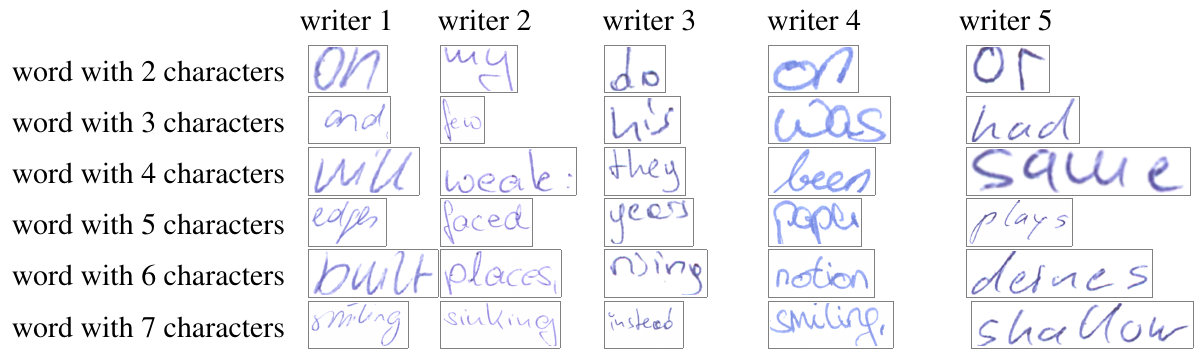}
	\caption{Examples of handwritten word images from the \textbf{CVL} dataset with different word lengths and writers. Each image has two attributes: word length and the {\color{blue}writer's identity}. }
	\label{fig:wordlen}
\end{figure}

Word length (number of letters in a word) is another visual attribute of handwritten word images. In this section, we {\color{blue}report on writer-identification experiments using} word length estimation as the auxiliary task. The maximum word length {\color{blue}for} both \textbf{CVL} and \textbf{IAM} is 13 {\color{blue}characters}. Therefore, the number of classes for word length estimation is 13. Fig.~\ref{fig:wordlen} shows an example of word images with different word lengths.

\begin{table}[!t]
	\centering 
	\caption{Performance of writer identification using different adaptive learning methods, with \textbf{word length estimation} as the auxiliary task on the \textbf{CVL} and \textbf{IAM} datasets.}
	\label{tab:wordlen}
	\resizebox{0.95\textwidth}{!}{
		\begin{tabular}{l|cc|cc|cc|cc}
			\hline\hline 
			\multirow{3}{*}{Model} & \multicolumn{4}{c|}{Writer Identification} & \multicolumn{4}{c}{Word Length Estimation (\textit{aux.})} \\
			\cline{2-9}
			 & \multicolumn{2}{c|}{\textbf{CVL}} & \multicolumn{2}{c|}{\textbf{IAM}} & \multicolumn{2}{c|}{\textbf{CVL}} & \multicolumn{2}{c}{\textbf{IAM}}\\
			\cline{2-9}
			& Top-1 & Top-5 & Top-1 & Top-5 & Top-1 & Top-5 & Top-1 & Top-5\\
			\hline 
			Baseline        & 75.3  & 92.5 & 66.0 & 82.9 & 94.3 & 99.9 & 91.5 & 99.8 \\
			Linear-adaptive & 75.9 & 92.7 & 65.4 & 83.1 & 92.7 & 99.9& 90.4 & 99.8\\
			Deep-adaptive    & \textbf{79.1} & \textbf{94.3} & \textbf{68.3} & \textbf{85.2} & 93.6 & 99.9 & 91.6 & 99.9\\
			\hline\hline
		\end{tabular}
	}
\end{table}

Table~\ref{tab:wordlen} shows the writer-identification performance based on single-word images with word length estimation as the auxiliary task.
From the table we can see that the word length is also an important attribute and transferring the learned features from word length estimation can also improve {\color{blue} writer-identification performance}.
{\color{blue}Like} the results in Table~\ref{tab:wordrec}, the \textbf{Deep-adaptive} model provides the best performance.

\begin{figure}[!t]
\centering
\includegraphics[width=0.95\textwidth]{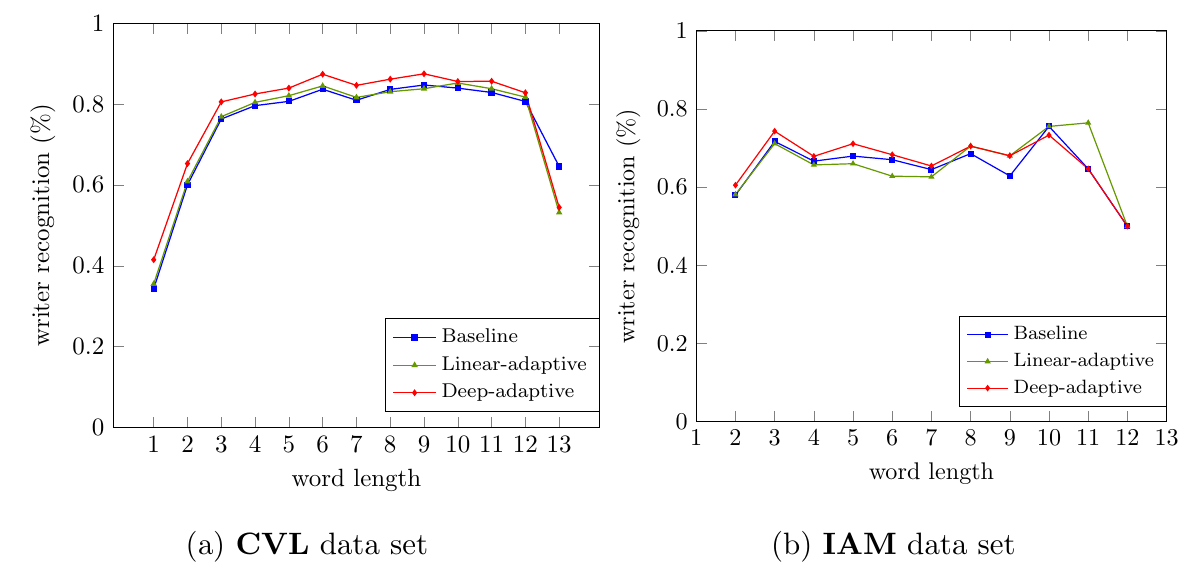}
\caption{Performance of writer identification (Top-1) {\color{blue}for} different CNN models with different word lengths on the \textbf{CVL} (Figure (a)) and \textbf{IAM} (Figure (b)) datasets.}
\label{fig:wordlenres}
\end{figure}

Fig.~\ref{fig:wordlenres} shows the writer-identification performance {\color{blue}for} different word lengths. From the figure, we can see that the performance of writer identification is much less sensitive {\color{blue}to} word length, unless {\color{blue}this} is greater than 2. 
{\color{blue}This could be because} word images with more than two characters contain more texts which can help to extract stable writing style information by deep learning. Another reason might be that resizing the word images with one or two characters introduces {\color{blue}more} noise than word images with more than two characters.
Note that the performance {\color{blue}for} word images {\color{blue}longer than eleven characters decreases} because there are few words with more than eleven characters on the \textbf{CVL} and \textbf{IAM} datasets, thus the
number of training samples is not sufficient.

\subsection{Performance of writer identification with character attribute recognition as auxiliary task}
Characters contained in the word are also important attributes and are used for word spotting in~\cite{almazan2014word,sudholt2016phocnet}.
In this section, we also {\color{blue}report on writer-identification experiments using} character attribute recognition as the auxiliary task.
We use similar attributes {\color{blue}to}~\cite{almazan2014word} and each word is represented by a binary histogram with 26 bins, corresponding to 26 English letters. Each element of this histogram represents whether the word {\color{blue}being studied contains the relevent letter}.
Note that we consider lower-case and upper-case letters as the same attribute because there are few upper-case letters in handwritten documents.
We also do not consider the spatial information {\color{blue}about} the characters in a word. 
For example, the word ``are" contains characters `a', `e' and `r', and their corresponding histogram bins are set to 1 and the others are zeros, the same as the PHOC histogram~\cite{almazan2014word} at the first level.
Character attribute recognition is a multiple-label learning problem. 
Therefore, we use the sigmoid activation function instead of softmax on the last layer of the auxiliary task.

Table~\ref{tab:wordattr} {\color{blue}presents the writer-identification performance} based on single-word images with character attribute recognition as the auxiliary task. From the table we obtain the same conclusion: the \textbf{Deep-adaptive} model {\color{blue}improves} the performance of writer identification {\color{blue}in both datasets}. 

\begin{figure*}[!t]
\centering 
\includegraphics[width=0.9\textwidth]{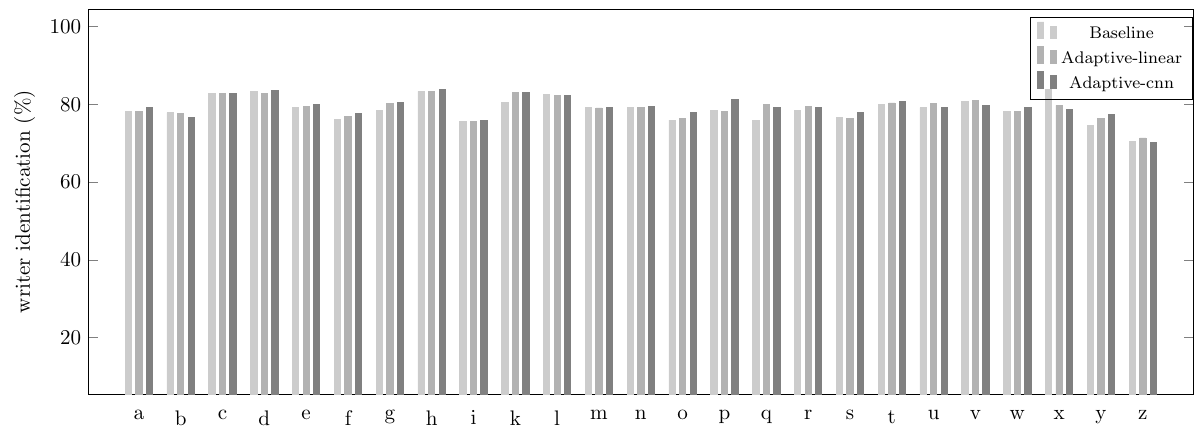}
\includegraphics[width=0.9\textwidth]{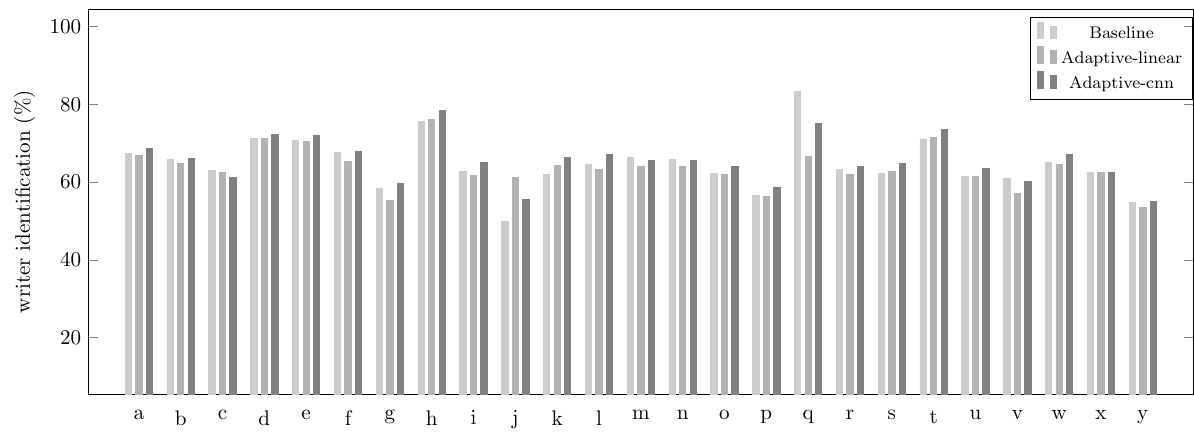}
\caption{The performance (Top-1) of writer identification with different character attributes. {\color{blue}The top figure shows} the performance {\color{blue}for} the \textbf{CVL} dataset {\color{blue}in which there is no word containing the} character `j' and the bottom figure shows the performance {\color{blue}for} the \textbf{IAM} dataset {\color{blue}in which there is no word containing the} character `z'.}
\label{fig:wordattr}
\end{figure*}

Fig.~\ref{fig:wordattr} shows the writer-identification performance of word images {\color{blue}containing} different characters.
From the figure we can {\color{blue}see} that all characters contain writing style information {\color{blue}about} the writer. The performance {\color{blue}for} word images which contain the characters `a',`d',`h',`t' is slightly higher than word images which contain other characters.
There are two possible reasons {\color{blue}for} different letters {\color{blue}containing} different amounts of handwriting style information:
(1) the shapes of {\color{blue}these} characters {\color{blue}are written differently} by different writers.
(2) These characters {\color{blue}typically touch others} in a cursive handwritten word and {\color{blue}the connecting shapes (ligatures) between the characters are also written differently by different writers}.

\begin{table}[!t]
	\centering 
	\caption{Performance of writer identification using deep adaptive learning with \textbf{character attribute recognition} as the auxiliary task on the \textbf{CVL} and \textbf{IAM} datasets.}
	\label{tab:wordattr}
	\resizebox{0.95\textwidth}{!}{
		\begin{tabular}{l|cc|cc|c|c}
			\hline\hline 
			\multirow{3}{*}{Model} & \multicolumn{4}{c|}{Writer Identification} & \multicolumn{2}{c}{Character Attribute Recognition (\textit{aux})} \\
			\cline{2-7}
			 & \multicolumn{2}{c|}{\textbf{CVL}} & \multicolumn{2}{c|}{\textbf{IAM}} & \textbf{CVL} & \textbf{IAM}\\
			\cline{2-7}
			& Top1 & Top5 & Top1 & Top5 & Accuracy &  Accuracy \\
			\hline 
			Baseline        & 75.1 & 92.6 & 65.9 & 83.4 & 93.4 & 91.3 \\
			Linear-adaptive & 75.3 &92.4  & 65.5 & 83.4 & 82.8 & 77.9 \\
			Deep-adaptive    & \textbf{76.5} & \textbf{93.2} & \textbf{67.6} & \textbf{84.3} & 85.1 & 81.6 \\
			
			\hline\hline
		\end{tabular}
	}
\end{table}

\subsection{Performance with reduced input image sizes}
In this section, we evaluate the writer-identification performance  to test the effect of reduced input image sizes.
A smaller input size {\color{blue}of} 32$\times$96$\times$1 was chosen to make sure that the {\color{blue}minimum} size of the last convolutional layer is greater than 1 pixel, since
there are four max-pooling layers with stride 2 in our network.
Tables~\ref{tab:diffsize_rec},~\ref{tab:diffsize_length}, and~\ref{tab:diffsize_encoding} show the writer-identification performance {\color{blue}for} different adaptive methods with different auxiliary tasks. 
From these tables we can see that the input size affects the writer-identification performance and {\color{blue}that} a smaller input size provides {\color{blue}poorer} results.
However, the {\color{blue}recognition performance} of the explicit information is approximately {\color{blue}the} same.
{\color{blue}This is because} recognition of the explicit information extracts whole-word characteristics, such as word shape and outline, which are less-sensitive to the word image size.
Conversely, the writer-identification model requires detailed features,
such as the curvature information of the ink traces, which are missing  or deformed in the small images.
It should be noted that the proposed \textbf{Deep-adaptive} model provides the best writer-identification performance {\color{blue}for} reduced image sizes, {\color{blue}albeit} less than when using large images with the same model.
Although training on large images takes more computing time, it provides better performance for writer identification (74.1\% vs 66.7\% average {\color{blue}of} \textbf{CVL} vs \textbf{IAM} with word recognition as the auxiliary task).
Therefore, we selected a 40$\times$120$\times$1 input size, which is a good trade-off between accuracy and efficiency.

\begin{table}[!t]
	\centering 
	\caption{Performance (Top-1) of writer identification with different input sizes, using different adaptive learning methods with \textbf{word recognition} as the auxiliary task on the \textbf{CVL} and \textbf{IAM} datasets.
	 W.I. means Writer Identification while  W.R. means Word Recognition in this table.}
	\label{tab:diffsize_rec}
	\resizebox{0.95\textwidth}{!}{
		\begin{tabular}{l|cc|cc|cc|cc}
			\hline\hline 
			\multirow{3}{*}{Model} & \multicolumn{4}{c|}{Input size: 40$\times$120$\times$1} & \multicolumn{4}{c}{Input size: 32$\times$96$\times$1} \\
			\cline{2-9}
			 & \multicolumn{2}{c|}{\textbf{CVL}} & \multicolumn{2}{c|}{\textbf{IAM}} & \multicolumn{2}{c|}{\textbf{CVL}} & \multicolumn{2}{c}{\textbf{IAM}}\\
			\cline{2-9}
			& W.I. & W.R. & W.I. & W.R. & W.I. & W.R. & W.I. & W.R.\\
					\hline 
		Baseline        & 75.3 & 95.1 & 65.7 & 93.5 & 66.7 & 95.1 & 61.6 & 94.2\\
		Linear-adaptive & 77.0 & 94.1 & 68.0 & 91.3 & 69.3 & 94.0 & 61.8 & 91.2\\
		Deep-adaptive    & 78.6 & 94.5 & 69.5 & 92.6 & 69.9 & 94.5 & 63.5 & 92.2\\
		\hline 
	Training Time  & \multicolumn{4}{c|}{8.5 hours} & \multicolumn{4}{c}{5.6 hours} \\
			\hline\hline
		\end{tabular}
	}
\end{table}

\begin{table}[!t]
	\centering 
	\caption{Performance (Top-1) of writer identification with different input sizes, using different adaptive learning methods with \textbf{word length estimation} as the auxiliary task on the \textbf{CVL} and \textbf{IAM} datasets.
	 W.I. means Writer Identification while  W.L.E. means Word Length Estimation in this table.}
	\label{tab:diffsize_length}
	\resizebox{0.95\textwidth}{!}{
		\begin{tabular}{l|cc|cc|cc|cc}
			\hline\hline 
			\multirow{3}{*}{Model} & \multicolumn{4}{c|}{Input size: 40$\times$120$\times$1} & \multicolumn{4}{c}{Input size: 32$\times$96$\times$1} \\
			\cline{2-9}
			 & \multicolumn{2}{c|}{\textbf{CVL}} & \multicolumn{2}{c|}{\textbf{IAM}} & \multicolumn{2}{c|}{\textbf{CVL}} & \multicolumn{2}{c}{\textbf{IAM}}\\
			\cline{2-9}
			& W.I. & W.L.E. & W.I. & W.L.E. & W.I. & W.L.E. & W.I. & W.L.E.\\
					\hline 
		Baseline        & 75.3 & 94.3 & 66.0 & 91.5 & 66.4 & 94.5 & 60.5 & 91.4\\
		Linear-adaptive & 75.9 & 92.7 & 65.4 & 90.4 & 68.4 & 92.8 & 59.2 & 89.4\\
		Deep-adaptive    & 79.1 & 93.6 & 68.3 & 91.6 & 69.9 & 93.6 & 61.8 & 90.2\\
		\hline 
	Training Time  & \multicolumn{4}{c|}{8.5 hours} & \multicolumn{4}{c}{5.6 hours} \\
			\hline\hline
		\end{tabular}
	}
\end{table}

\begin{table}[!t]
	\centering 
	\caption{Performance (Top-1) of writer identification with different input sizes, using different adaptive learning methods with \textbf{word attribute recognition} as the auxiliary task on the \textbf{CVL} and \textbf{IAM} datasets.
	 W.I. means Writer Identification while  W.A.R. means Word Attribute Recognition in this table.}
	\label{tab:diffsize_encoding}
	\resizebox{0.95\textwidth}{!}{
		\begin{tabular}{l|cc|cc|cc|cc}
			\hline\hline 
			\multirow{3}{*}{Model} & \multicolumn{4}{c|}{Input size: 40$\times$120$\times$1} & \multicolumn{4}{c}{Input size: 32$\times$96$\times$1} \\
			\cline{2-9}
			 & \multicolumn{2}{c|}{\textbf{CVL}} & \multicolumn{2}{c|}{\textbf{IAM}} & \multicolumn{2}{c|}{\textbf{CVL}} & \multicolumn{2}{c}{\textbf{IAM}}\\
			\cline{2-9}
			& W.I. & W.A.R. & W.I. & W.A.R. & W.I. & W.A.R. & W.I. & W.A.R.\\
					\hline 
		Baseline         & 75.1 & 93.4 & 65.9 & 91.3 & 67.6 & 93.6 & 60.1 & 90.6\\
		Linear-adaptive  & 75.3 & 82.8 & 65.5 & 77.9 & 69.7 & 83.9 & 60.6 & 76.8\\
		Deep-adaptive    & 76.5 & 85.1 & 67.6 & 81.6 & 70.4 & 86.1 & 63.5 & 82.3\\
		\hline 
	Training Time  & \multicolumn{4}{c|}{8.5 hours} & \multicolumn{4}{c}{5.6 hours} \\
			\hline\hline
		\end{tabular}
	}
\end{table}

\subsection{Comparison with other studies}   
{\color{blue}This section compares other writer identification methods using the} \textbf{CVL} and \textbf{IAM} datasets based on single-word images.
For the handcrafted features, we used the ``leave-one-out" strategy, the same as the traditional writer identification approach~\cite{bulacu2007text,siddiqi2010text}.
The representation of each writer is computed as the average word features except the query one.
Table~\ref{tab:soa} shows the performance of the different writer-identification methods.
From the table, we can see that the traditional handcrafted features fail {\color{blue}to identify the writer} based on single-word images, which is also shown in~\cite{brink2008much}.
The CNN model provides much better results than the handcrafted features, and our proposed deep adaptive learning method provides the best results.

\begin{table}[!h]
	\centering
	\caption{Single-word writer-identification performance {\color{blue}using} different approaches on the \textbf{CVL} and \textbf{IAM} datasets. }
	\label{tab:soa}
	\begin{tabular}{l|cc|cc}
	\hline\hline 
	\multirow{2}{*}{Method } &    \multicolumn{2}{c|}{\textbf{CVL}} &   \multicolumn{2}{c}{\textbf{IAM}} \\
	\cline{2-5}
                                                	& Top1 & Top5 	& Top1 & Top5 \\
	\hline 
	Hinge~\cite{bulacu2007text}                     & 25.8  & 48.0 & 26.7 & 45.4 \\
        Quill~\cite{brink2012writer}	                & 29.4  & 52.6 & 35.9 & 57.8 \\
	Chain Code Pairs~\cite{siddiqi2010text}	        & 22.4  & 44.6 & 21.6 & 39.7 \\
	Chain Code Triplets~\cite{siddiqi2010text}      & 28.8  & 51.4 & 30.5 & 49.8 \\
	COLD~\cite{he2017writer}		        & 12.8  & 29.6 & 15.7 & 32.1\\	
	QuadHinge~\cite{he2016co}		        & 30.0  & 52.4 & 37.2 & 57.8 \\
	CoHinge~\cite{he2016co} 		        &  25.9 & 46.9 & 26.8 & 47.2 \\
	\hline
	CNN~\cite{NIPS2012_4824}                        & 75.3  & 92.6 & 66.0 & 83.5 \\
	CNN+Adaptive                                    & 79.1  & 93.7 & 69.5 & 86.1\\

	\hline\hline 
	\end{tabular}
\end{table}

\subsection{Discussion}

From {\color{blue}Tables}~\ref{tab:wordrec}, \ref{tab:wordlen} and \ref{tab:wordattr}, we can see the following.
(1) Generally, {\color{blue}other conditions being equal}, recognizing implicit information (writer identification) is more difficult than {\color{blue}recognizing} explicit information such as word recognition, word length estimation and character attribute recognition.
Since the implicit information is embedded in {\color{blue}the} patterns of handwritten characters or ink traces, it usually needs more reference data to be recognized correctly.
(2) Adaptive learning can improve the performance of the main task. For example, the writer identification performance of the \textbf{Linear-adaptive} and \textbf{Deep-adaptive} models with three different auxiliary tasks is better than {\color{blue}that of} the \textbf{Baseline} model on both two datasets.
(3) The writer identification performance of the \textbf{Deep-adaptive} model is better than {\color{blue}that of} the \textbf{Linear-adaptive} model. This is because the deep adaptive learning model can learn the non-linear relationship between different tasks.
(4) The performance of the auxiliary task decreases {\color{blue}in} adaptive learning because the {\color{blue}main task information} is back-propagated to the auxiliary task layers. However, the  \textbf{Deep-adaptive} performance is better than {\color{blue}that of} the \textbf{Linear-adaptive} model, {\color{blue}showing} that the residual adaptive blocks $C(\cdot)$ can transfer the useful information from the auxiliary task to the main task on the forward phase and mask the useless information back-propagated to the auxiliary task.
(5) Using word recognition and word length estimation as the auxiliary tasks {\color{blue}yields} better results for writer identification {\color{blue}in} the two datasets than using character attribute recognition (see Table~\ref{tab:compr}).
{\color{blue}This could be because} the character attribute recognition results {\color{blue}are} not a good choice as an auxiliary task, thus the learned features contain less useful information.
Therefore, choosing {\color{blue}a high performing auxiliary task can also result in a greater improvement in the main task}.
(6) We also {\color{blue}attempted to combine} all three auxiliary tasks together in our experiments, considering the word {\color{blue}itself} and word length as attributes, similar {\color{blue}to the} character attributes. 
The results are shown in Table~\ref{tab:compr} and we can see that combining all {\color{blue}the} auxiliary tasks cannot improve performance. {\color{blue}This could be because} during training, the loss is dominated by the character attributes.
For example, the word ``Imagine" has 7 character bits and only 1 word bit and 1 word length bit. Thus, the neural network focuses on recognizing the character attributes, which {\color{blue}results in a poorer} performance than {\color{blue}that of the} other two auxiliary tasks.
(7) The large performance difference between traditional methods and CNN {\color{blue}for} writer identification based on difficult single-word images (see Table~\ref{tab:soa}) indicates that  the necessary information for writer identification is somehow present in individual words. 
However, as with most CNN methods, there may be some over-fitting {\color{blue}which led} to the current results. 
More research is needed to assess the effectiveness of the dropout mechanism used during training, {\color{blue}for instance}.

\begin{table}
	\centering
	\caption{Final overview of the writer-identification performance using the \textbf{Deep-adaptive} model with different auxiliary tasks.}
	\label{tab:compr}
	\resizebox{0.95\textwidth}{!}{
	\begin{tabular}{l|cc|cc}
		\hline\hline 
		\multirow{2}{*}{Auxiliary Tasks} & \multicolumn{2}{c|}{\textbf{CVL}} & \multicolumn{2}{c}{\textbf{IAM}} \\
		\cline{2-5}
		 & Top1 & Top5 & Top1 & Top5\\
		\hline 
		Baseline & 75.2 & 92.5 & 65.8 & 83.3 \\
		\hline 
		Word Recognition & 78.6 & 93.7 & \textbf{69.5} & \textbf{86.1} \\
		Word Length Recognition & \textbf{79.1} & \textbf{94.3} & 68.3 & 85.2 \\
		Character Attribute Recognition & 76.5 & 93.2 & 67.6 & 84.3\\
		Combined  & 78.5 & 94.0 & 67.5 & 84.3 \\
		\hline\hline
	\end{tabular}
}
\end{table}

The experimental results provide several interesting factors {\color{blue}to consider when} designing a modern writer identification system: 
(1) it is better to ask writers to write more words, with at least five words to achieve a high performance.
(2) Since the writer identification performance of word images with less than two characters is very low, it is better to ask writers to write words with as least three characters and each word should contain writing-sensitive letters, such as `a', `d', `h', and `t'.

\section{Conclusion}
\label{sec:conclusion}

This paper has studied the writer identification problem based on single-word images using deep adaptive learning in a multi-task learning framework.
Three different tasks which recognize the explicit information of handwritten word images {\color{blue}were} used as the auxiliary tasks to improve the performance of writer identification.
{\color{blue}The} experimental results on two benchmark datasets have shown several interesting conclusions.
Firstly, writer identification is more difficult than other attribute recognition problems because the {\color{blue}writer's identity} is the implicit information, and {\color{blue}even people themselves find recognizing a writer based on single-word images difficult}.
Secondly, adaptive learning can improve the performance of writer identification since different tasks learn different features and the specific representations of the auxiliary task can be transferred to the main task.
Thirdly, deep adaptive learning can capture the complex relationship between the specific features of different tasks and can thus provide better performance.

The performance of writer identification based on single-word images is still {\color{blue}much poorer} compared to the performance of other tasks using deep learning, and it {\color{blue}still needs} to be improved in the future.
Recently, there {\color{blue}has been} a big shift from handcrafted features to handcrafted structures {\color{blue}in} neural networks.  Therefore, more complex neural network structures can be investigated in the future for writer identification.


\section*{References}

\bibliography{transfernet}

\end{document}